\documentclass{article}

\usepackage[nonatbib,final]{nips_2017}
\usepackage{amsmath}
\usepackage{graphicx}

\usepackage[utf8]{inputenc} % allow utf-8 input
\usepackage[T1]{fontenc}    % use 8-bit T1 fonts
\usepackage{hyperref}       % hyperlinks
\usepackage{url}            % simple URL typesetting
\usepackage{booktabs}       % professional-quality tables
\usepackage{amsfonts}       % blackboard math symbols
\usepackage{microtype}      % microtypography

\title{Speedup from a different parametrization within the Neural Network algorithm}

\author{
  Michael F.~Zimmer \\
  Neomath, Inc.  \\
  \texttt{mzimmer@neomath.com} \\
}

\begin{document}
% \nipsfinalcopy is no longer used

\maketitle

\begin{abstract}
A different parametrization of the hyperplanes is used in the neural network algorithm.
As demonstrated on several autoencoder examples it significantly outperforms the usual parametrization, reaching lower training error values with only a fraction of the number of epochs.
It's argued that it makes it easier to understand and initialize the parameters.
\end{abstract}

\section{Introduction}

Artificial Neural Networks (or Neural Nets (NN)) are a powerful and versatile machine learning algorithm.
They act as a universal function approximator, and can be used in classification and regression problems\cite{goodfellow,bishop,rojas}.
However, despite their successes, a problem that persists is long training times of the NN.
This paper addresses that problem by proposing a different parametrization of the hyperplanes used in the algorithm.

This paper demonstrates the new approach on five autoencoder examples.
It allows the training to proceed much more rapidly, using in some cases only 1/8-th the number of epochs of the existing approach to reach a given training error value.
Results are given for individual runs, averaged runs, and the "epoch speedup".
Also, this different parametrization is more intuitive, making it easier to understand how the parameters work,
and thus provides insight into how one might best initialize a NN.

\section{NN definitions}

\begin{figure}
  \centering
  \includegraphics[scale=0.8]{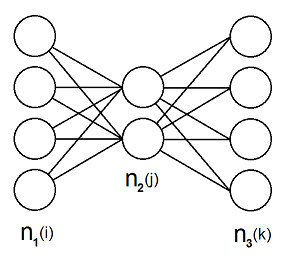}
  \caption{Autoencoder with a 4-2-4 network.}
  \label{fig:autoenc}
\end{figure}

Here the architecture and variables are defined for the 3-layer NN that will be used in the examples.
It is a feed-forward NN with $d_1$, $d_2$, $d_3$ nodes on each of the three layers; they are labeled as $n_1(i)$, $n_2(j)$, $n_3(k)$, 
with indices $i=1,..,d_1$, $j=1,...,d_2$, and $k=1,...,d_3$.
In Fig.~\ref{fig:autoenc} is shown an example where $d_1=4$, $d_2=2$, and $d_3=4$.
For example, the transformations on $n_1$ which lead to $n_2$ are defined as:
\begin{align*}
	n_2(j) & = f(h_1(j))  \\
	h_1(j) & = w_1(j,0) + \sum_{1,d2} w_1(j,i) n_1(i)  \\
	f(h)   & = 2 / (1 + e^{-h}) - 1
\end{align*}
Note that this choice of $f$ enforces a $(-1,1)$ encoding, as opposed to the $(0,1)$ encoding; this will be discussed later.
The training error ($E$) is a mean-squared error; a classification error could also be used.
\begin{equation*}
	E = \sum_k [ n_3(k) - t(k) ]^2
\end{equation*}
Back-propagation is used, which is based on differential updates made to $w = (w_0,w_1,...,w_d)$ using: 
\begin{equation*}
	\Delta w = -\eta \frac{\partial E}{\partial w}
\end{equation*}
where $\eta$ is the {\it learning rate}.

\section{Different Parameterization}

It may be the reader's experience, as it has been the author's, that there are advantages to having a simple parametrization of the system being optimized.  
In particular, it's beneficial to have a 1-to-1 mapping between parameter values with the actual configuration of your system (i.e., the hyperplane configurations); that is not the case here.
E.g., all $w$ values could be scaled up and down, and they would all describe the same hyperplane $w \cdot x=0$; that scale change is important for the function $f$, however.
Another point is the lack of intuition that the usual parametrization provides: it's simply not clear by looking at the parameter values, without any additional calculation,
whether the w-values are appropriate for a given data set.
Toward finding a different parametrization that doesn't suffer from these shortcomings is the motivation for this paper.

The key point about the different parametrization is that it is based on features of the hyperplane-- i.e., the minimum distance
between the origin and the hyperplane ($c$), and the orientation of the hyperplane ($u$).  
Using those variables a hyperplane ($L$) can be written as $L = c + u \cdot x$, where $x$ is a point in space (i.e., the data).
While clearly it's almost the same as what has been traditionally used, it has the advantage that it's a unique representation of the hyperplane.  It can easily be checked that for each $(c,u)$ combination, there is exactly one hyperplane, and vice versa.  Also, it is easy to partition all possible hyperplanes, by first fixing $c$ or $u$.

The two parametrizations are trivially connected (using $x_0 = 1$); the difference is having the overall scale pulled out:
\begin{align*}
	h & = w_0 + w_1 x_1 + ... + w_d  x_d  \\
	  & = s ( c + u \cdot x )
\end{align*}
where $w' = (w_1, ..., w_d)$, $s = ||w'||$, $c = w_0/||w'||$, $u_i = w_i / ||w'||$, $i=1,...,d$, and $|| \cdot ||$
denotes the $L_2$ norm.
For simplicity, the new parameters are combined into one as $z=(s,c,u)$.  
The new parametrization will be referred to as {\bf z-param}, and the traditional one as {\bf w-param}.
Optimizations for z-param will be done against the scalars $s$,$c$ and the vector $u$.

\section{Examples of New Approach}

A comparison is now made between the two parametrizations on a clean but sufficiently challenging test case: an autoencoder.
Five 3-layer autoencoder examples, ranging in size from 8-3-8 to 128-7-128, are used to compare w-param and z-param.
The sizes $(d_1,d_2,d_3)$ were chosen to satisfy $d_3=d_1$ and $d_2=log_2(d_1)$.
This architecture is chosen to force the NN to learn a compact representation of the $d_1$ data points using only $d_2$ variables.
For example, in Mitchell\cite{mitchell} it is explained how the data $(1,0,0,0,0,0,0,0)$, $(0,1,0,0,0,0,0,0)$,... essentially becomes encoded by $(1,0,0)$, $(0,1,1)$,..., 
where the $d_2$ layer is reminiscent of a binary encoding.  
In this supervised task, the target data ($t$) forces the nodes ($n_3$) on the third layer to reproduce the input data.
Also, to keep this comparison as straightforward as possible, certain techniques were left out that are often employed: i.e., momentum, adjustable decay rate, mini-batch.
Were these features to be included, they could be applied to both w-param and z-param.

Code for running these examples is available on a public repository\cite{github}.  Pseudocode will not be shown, since it is the same as the existing NN algorithm, and differs only with respect to the parameter updating.  Here $\Delta w = -\eta \partial E/\partial w$ is replaced by its z-equivalent, and since $z=(s,c,u)$ there are three separate derivatives for each layer.  This leads to the 6 derivatives shown below, which will be used for updating those parameters via $\Delta z = -\eta  \partial E/\partial z$,
\begin{align*}
	\partial E/\partial s_2(p)   & = \left[c_2(p) + \sum_j u_2(p,j) n_2(j) \right] A(p)  \\
	\partial E/\partial c_2(p)   & = s_2(p) A(p)  \\
	\partial E/\partial u_2(p,q) & = s_2(p) n_2(q) A(p)  \\
	\partial E/\partial s_1(p)   & = \left[ c_1(p) + \sum_i u_1(p,i) n_1(i) \right] B(p)  \\
	\partial E/\partial c_1(p)   & = s_1(p) B(p)  \\
	\partial E/\partial u_1(p,q) & = s_1(p) n_1(q) B(p)
\end{align*}
where 
\begin{align*}
	A(p)    & = f'(h_2(p)) [n_3(p) - t(p) ]  \\
	B(p)    & = f'(h_1(p)) \sum_m s_2(m) u_2(m,p) A(m)  \\
	h_v(p)  & = s_v(p) \left[ c_v(p) + \sum_i u_v(p,q) n_v(q) \right]  
\end{align*}
and $v=1,2$.
Noticeable is that there is nothing in the model that enforces $||u||=1$.  It is skipped in this presentation and will be discussed at a future time.
The initial values for $z = (s,c,u)$ are: $s=1$, $c=0$, $u$ = random unit vector.  Note that the same initialization is used for both w-param and z-param;
the difference in the results follows from the different updating schemes applied to each.
The choice of the initialization value $s=1$ was made to simplify the comparison between these two algorithms.  In fact, using a very small non-zero $s$ leads to better results,
and will be explored briefly later in this section.

Before comparing the two approaches over those different examples, an (approximate) optimal $\eta$ will be determined for each.  This was done using a grid search over $\eta$, averaging 20 runs for each point (10 runs for $d_1=128$).  The results for the best $\eta$ for z-param (i.e., $\eta_z$) and for w-param (i.e., $\eta_w$) are in Table~\ref{tbl_eta}.

\begin{figure}[!htb]
  \centering
  \begin{minipage}[b]{0.49\textwidth}
    \includegraphics[width=\textwidth]{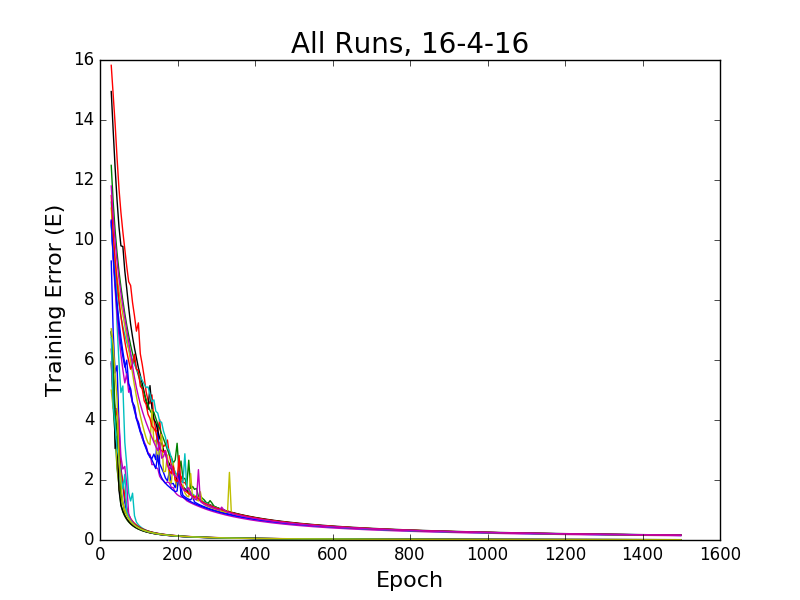}
    \caption{z-param is the lower grouping.}
    \label{fig:group16}
  \end{minipage}
  \hfill
  \begin{minipage}[b]{0.49\textwidth}
    \includegraphics[width=\textwidth]{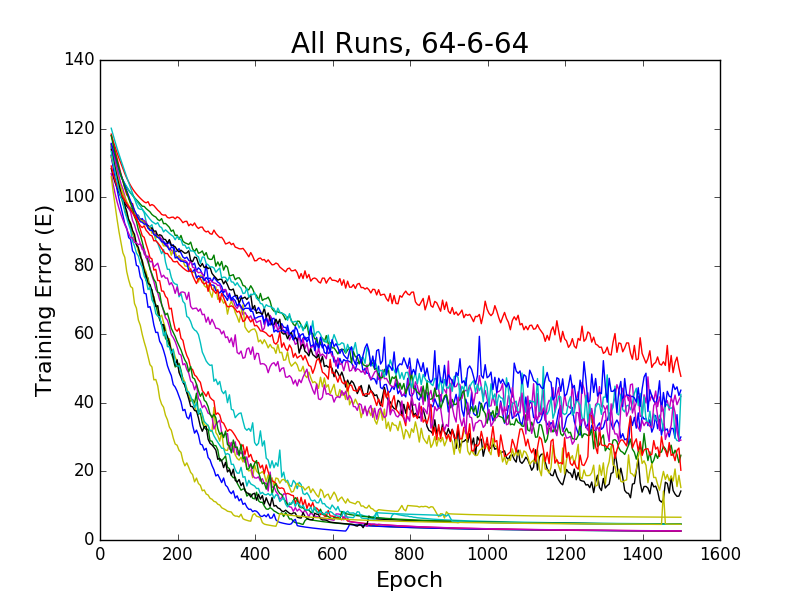}
    \caption{z-param is the lower grouping.}
    \label{fig:group64}
  \end{minipage}
\end{figure}

\begin{table}[t]
  \caption{$\eta$ values used}
  \label{tbl_eta}
  \centering
  \begin{tabular}{lll}
    \toprule
	$d_1$ &  $\eta_z$ & $\eta_w$   \\
    \midrule
	8    &  0.180    &  0.45     \\
	16   &  0.110    &  0.22     \\
	32   &  0.040    &  0.12    \\
	64   &  0.016   &  0.07     \\
	128  &  0.010    &  0.06    \\
    \bottomrule
  \end{tabular}
\end{table}
Using the $\eta_w$ and $\eta_z$, the w-param and z-param algorithms were run over 1500 epochs, starting from the same initial conditions (as described earlier).  The results are overlaid on each other: one plot for $d_1=16$ (Fig.~\ref{fig:group16}) and one for $d_1=64$ (Fig.~\ref{fig:group64}). 
Note that in both graphs the better-performing, lower grouping corresponds to z-param, the new parametrization.
These results were typical of the entire set: the z-param always performed better, reaching a lower training error more quickly.
Most noticeable is that as $d_1$ increases, the variance of the final $E$ increases, as well as the final $E$ value itself.

\begin{figure}[!tbp]
  \centering
  \begin{minipage}[b]{0.49\textwidth}
    \includegraphics[width=\textwidth]{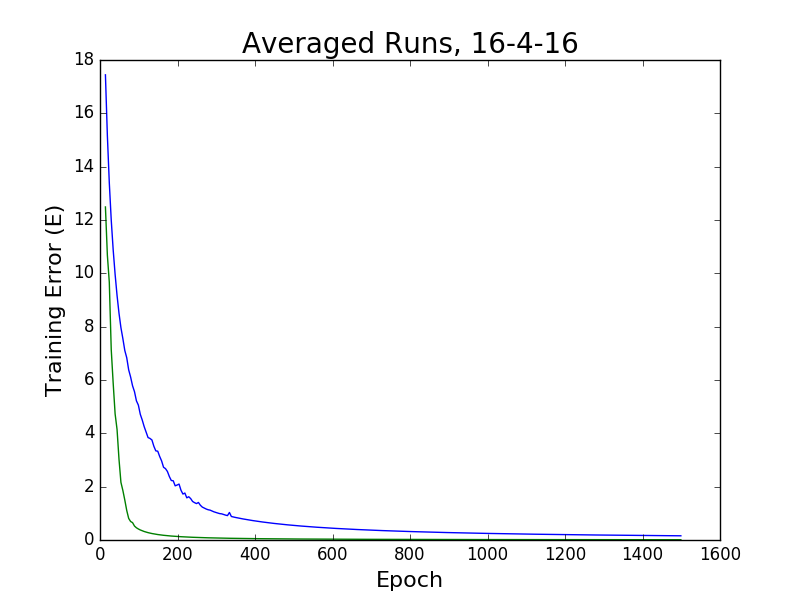}
    \caption{z-param is the lower curve.}
    \label{fig:avg16}
  \end{minipage}
  \hfill
  \begin{minipage}[b]{0.49\textwidth}
    \includegraphics[width=\textwidth]{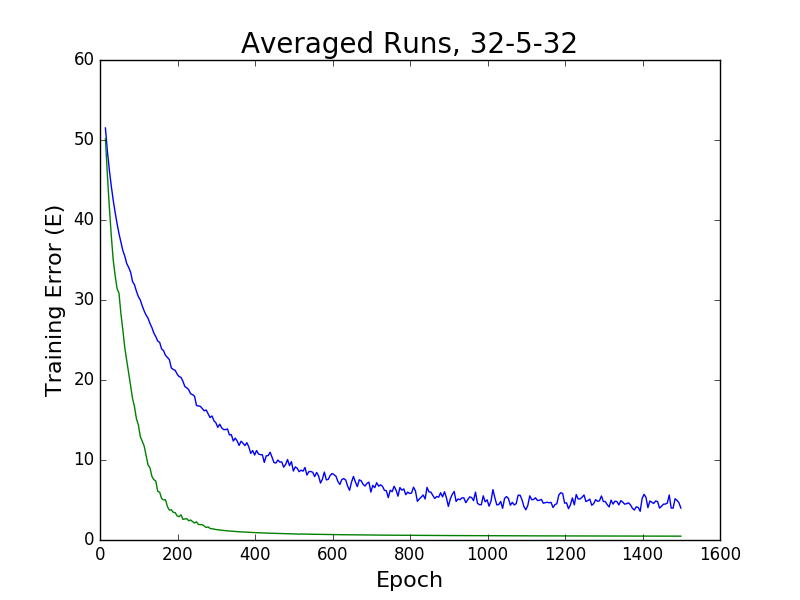}
    \caption{z-param is the lower curve.}
    \label{fig:avg32}
  \end{minipage}
\end{figure}

\begin{figure}[!tbp]
  \centering
  \begin{minipage}[b]{0.49\textwidth}
    \includegraphics[width=\textwidth]{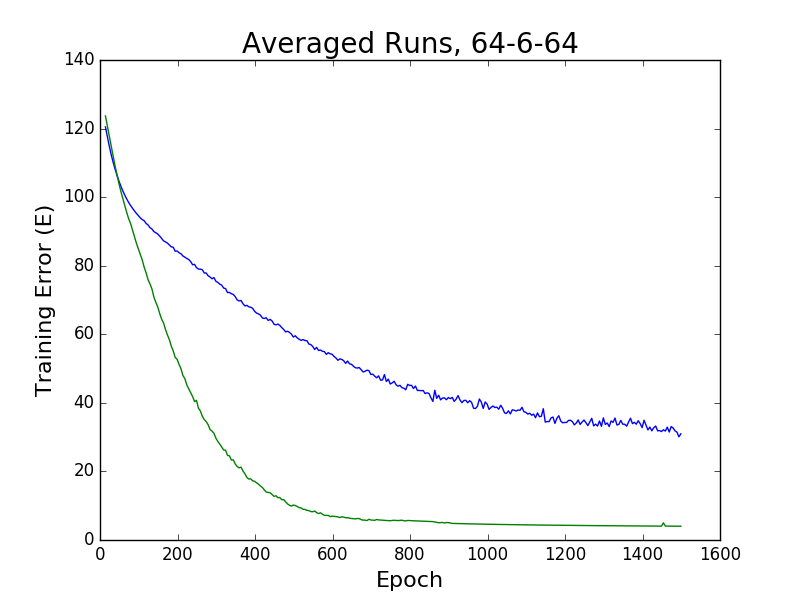}
    \caption{z-param is the lower curve.}
    \label{fig:avg64}
  \end{minipage}
  \hfill
  \begin{minipage}[b]{0.49\textwidth}
    \includegraphics[width=\textwidth]{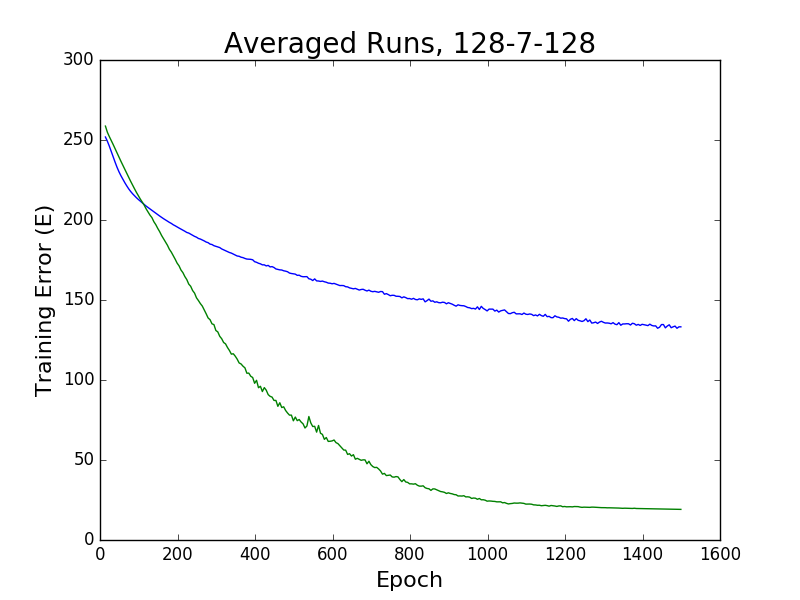}
    \caption{z-param is the lower curve.}
    \label{fig:avg128}
  \end{minipage}
\end{figure}

To facilitate seeing the general trend as $d_1$ increases, those runs are averaged for each $d_1$ to produce Fig.~\ref{fig:avg16} through Fig.~\ref{fig:avg128}.  This is displayed in the figures for $d_1=16,32,64,128$.
These graphs make it more obvious that z-param always outperforms w-param.  Indeed, the final $E$ of the z-param curves are typically
about 1/10-th that of the w-param.
It bears repeating that these two cases began from
the same initial conditions on the hyperplanes and used the same mean squared error; the only difference is that the hyperplanes were parametrized and updated differently.

\begin{figure}
  \includegraphics[scale=0.65]{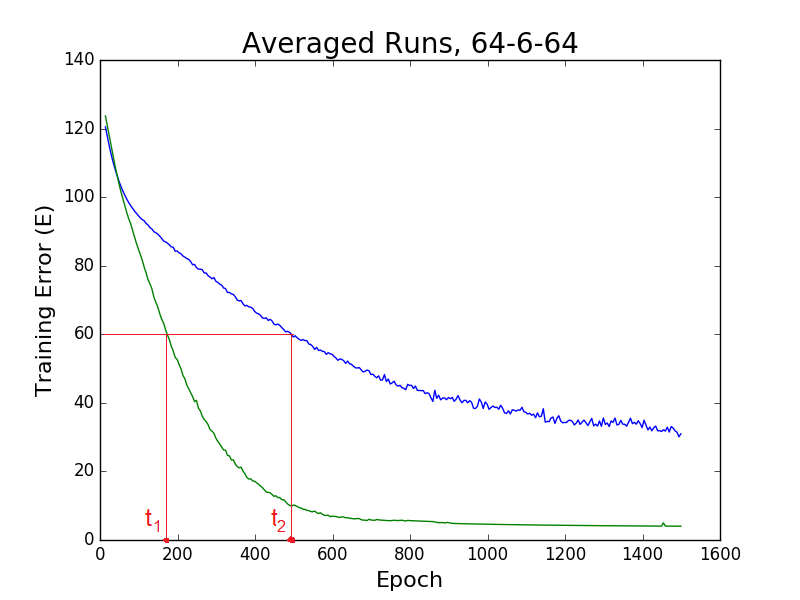}
  \caption{The speedup is $t_2$ / $t_1$.}
  \label{fig:epoch_t2t1}
\end{figure}

To better quantify the advantage z-param has over w-param, an "epoch speedup" is visually defined in Fig.~\ref{fig:epoch_t2t1} with respect to the averaged runs (cf. Fig.~\ref{fig:avg16}); it's simply the ratio of epochs ($t_2/t_1$) needed to reach a given $E$\cite{smoothing}.
Note this can only be computed at a training error ($E$) where both curves have a value.
E.g., for very low training errors, only z-param has values, and the speedup cannot be computed.
Thus the window of training error ranges from the min of the max $E$ values to the max of the min $E$ values, for each pair of w and z training error values.
That range is then scaled from 0 to 100, and is referred to as the "Percent toward 0".

\begin{figure}
  \includegraphics[scale=0.65]{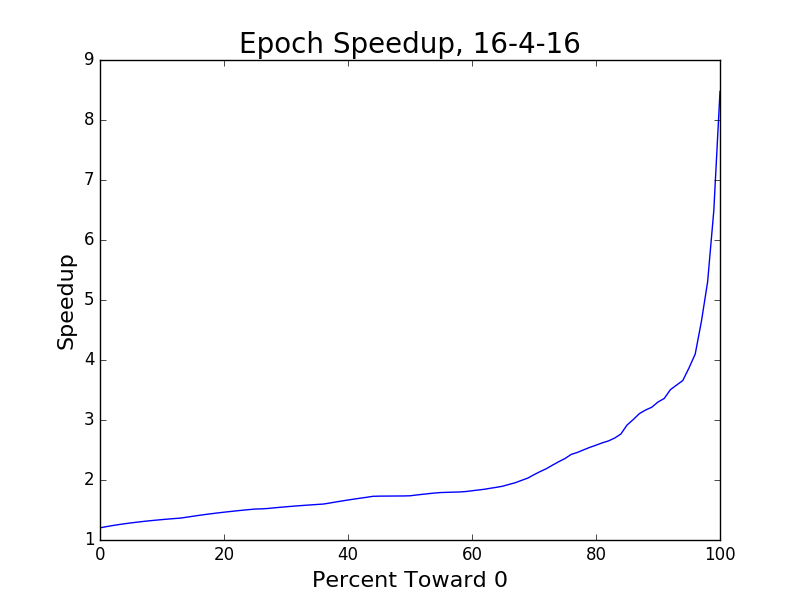}
  \caption{Typical epoch speedup.}
  \label{fig:speedup}
\end{figure}

Shown in Fig.~\ref{fig:speedup} is a typical graph of the epoch speedup, in this case for $d_1=16$.
Of note is how the speedup increases as the original curves reached lower $E$ values.  There is no reason to believe this trend would not continue for higher epoch number.
Also, in a sense these graphs don't tell the whole story-- the w-param appears to not be able to reach (in a reasonable time) the low $E$ values of the z-param.

The maximum speedup recorded for all the cases is shown in Table~\ref{tbl_speedup}
\begin{table}[t]
  \caption{Epoch speedups}
  \label{tbl_speedup}
  \centering
  \begin{tabular}{ll}
    \toprule
    $d_1$ 	&  speedup     \\
    \midrule
    8 	& 6.75       \\
    16 	& 8.5       \\
    32 	& 7.75       \\
    64 	& 5.1       \\
    128 & 5.0       \\
    \bottomrule
  \end{tabular}
\end{table}

\subsection*{Small s}

\begin{figure}
  \includegraphics[scale=0.65]{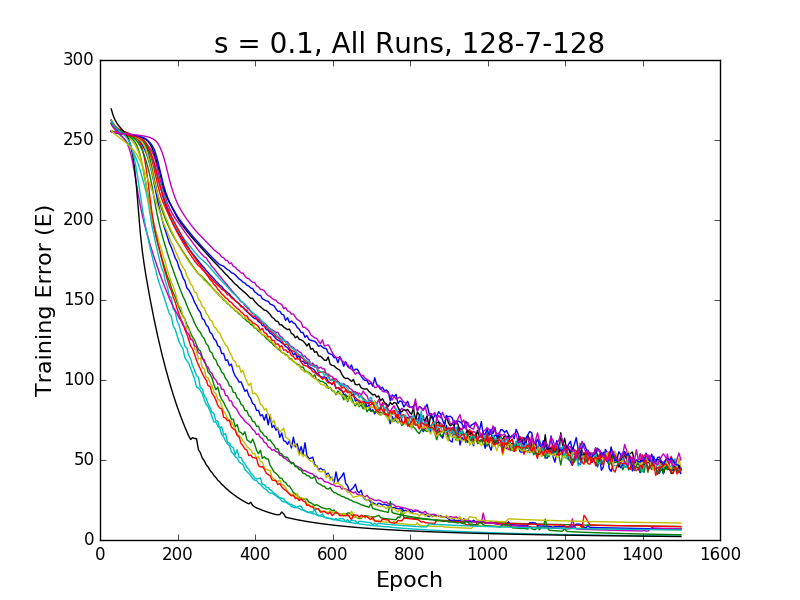}
  \caption{Lower grouping is z-param.}
  \label{fig:group128_0.1}
\end{figure}

\begin{figure}
  \includegraphics[scale=0.65]{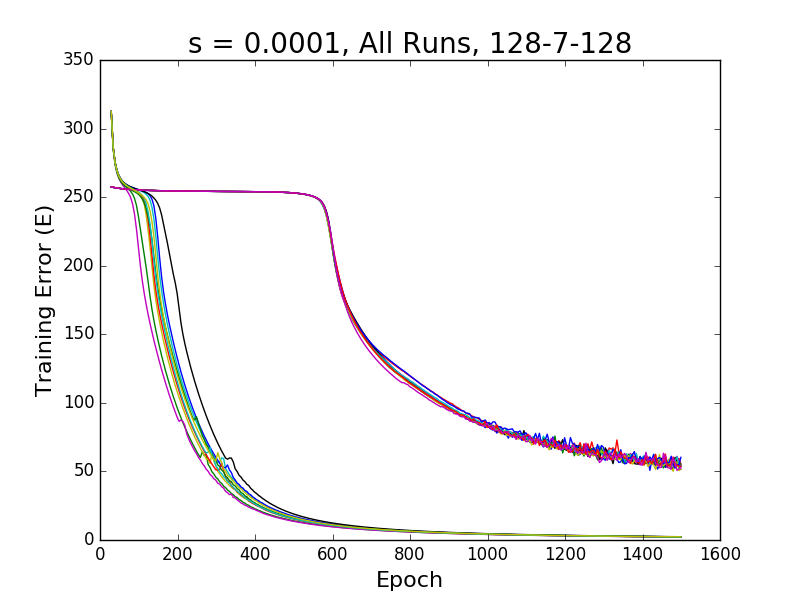}
  \caption{Lower grouping is z-param.}
  \label{fig:group128_0.0001}
\end{figure}

The above comparisons were all made using the initial condition of $s=1$, which was done in large part to simplify the comparison.
Having done that, one may now explore the effects of small $s$.  Shown in Fig.~\ref{fig:group128_0.1} are the w-param and z-param for $d_1=128$ for $s=0.1$
Again z-param does better, but what is notable is how both outperformed their corresponding $s=1.0$ versions.  Qualitatively, one can see the different runs coming together, and having a smaller variance for larger epoch values.  There is also the emergence of a plateau early on, more easily seen in the w-param case.  To explore this still further, the same graph was produced in Fig.~\ref{fig:group128_0.0001}, except using $s=0.0001$
The quality of the z-param (lower curves) is evident: it's shifted left, reaches lower values, and has reduced variance.  The w-param graphs have an increased plateau and less variance, but don't reach noticeably lower $E$ values.

\section{Parameter Initialization}

This paper, while begun as an investigation into a different hyperplane parametrization, has also shed light on optimal parameter initializations.
Beginning from the parameters for the z-param case, there are three types of parameters: the scale of $h$ (i.e., $s$), the distance of hyperplane to origin ($c$), and the orientation of hyperplane ($u$).
	
The scale ($s$) was best chosen to be a small number.  In Fig.~\ref{fig:group128_0.0001} it was seen to lead to a plateau region initially (especially for the w-param case).
Presumably this allowed more time for $c$ and $u$ parameters to find more optimal values, thereby negating the influence of an initial random $u$.
This would happen because, since $f(h) = f(s(c+u.x))$, a small $s$ would hide modest changes in $c$ and $u$, thereby initially allowing those parameters to more easily evolve toward optimal values.
There is evidence for this in that the curves have coalesced for $s=0.1$ in Fig.~\ref{fig:group128_0.1}, and have strongly coalesced as $s$ was made smaller to $s=0.0001$ in Fig.~\ref{fig:group128_0.0001}.
Thus, when using z-param, it seems best to set $0 < s \ll 1$, at least when there is a significant variance in individual runs as was seen earlier.  Likewise, for the w-param case, the scale is set by $||w'||$, where again $w' = (w_1, ..., w_d)$,
so the corresponding scaled values are:
\begin{equation*}
	w_i \rightarrow (s/||w'||) w_i
\end{equation*}
where $s$ is the same as for z-param.

The distance of a hyperplane to origin ($c$) was set to 0, under the assumption that the data was shifted about its mean.
This would ensure that all the data points do not fall on one side of the plane, making the hyperplane much less useful\cite{rojas}.
The effect of having a hyperplane completely on one side of a data cloud is that $h$ will become very large in magnitude, 
which leads to $f(h)$ being pinned to its min/max values, and also $f'(h)$ being near zero.  
This is known as the zero-gradient problem.  
Here it is called being "caught in the doldrums", a sailing term used to describe a situation near the equator, where there is no wind, and hence no prospect of making progress.
Being caught in the doldrums is equally disadvantageous for training a NN.  To avoid this scenario, and ensure the hyperplanes (nearly) always intersect the data cloud, the input layer is shifted about its mean.  For later layers, it's preferable to not have to keep doing that, so to mitigate that the $(-1,1)$ encoding is used
instead of the $(0,1)$ encoding.  This has the advantage of having positive and negative values, and hence being somewhat self-averaging (toward 0).

The orientation of the hyperplane ($u$) should obviously be in a random direction.  A correct prescription for constructing
such a vector is to draw from a {\it normal} distribution for each component, and then normalize the result\cite{randomvector}.
If one instead draws from a uniform distribution (as is commonly done), and normalizes that, the resulting $u$ will be biased toward the corners of the enclosing hypercube.
The same approach can be used with $w'$ (in w-param) as for $u$ (in z-param).

\section{Final Remarks}

In this paper, the ramifications of a different parametrization (z-param) of the hyperplanes were investigated in the NN algorithm.
The motivation was that a more intuitive and straightforward representation could help the hyperplanes more easily explore their parameter space.

The results show that by switching to z-param, learning can be achieved much more quickly.
It's evident from the "epoch speedup" graphs (cf. Fig.~\ref{fig:speedup}) that as training continues, there is increasingly more to be gained from using the z-param.
Keep in mind though that the epoch speedup numbers are not perfectly representative of the wall clock time for implementing this algorithm.
The FLOPS are slightly higher for z-param, and the parallelizability may be different (although it appears similar).

In addition to the above, the z-param approach presented a simpler situation with respect to initializing the parameters.  In the process it was noted, at least for the examples studied, that high variance training error runs might be a motivation for instead initializing with a very small $s$.

Investigations into this new parametrization have just begun and 
there are clearly many avenues of research left to be pursued.
Also, since this is only a change in the parametrization of the hyperplanes, it means it can be substituted
into existing NN applications that use the existing w-param.  Hence, this new parametrization can be applied on a wide scale.


\begin{thebibliography}{10}
\bibitem{smoothing}
Additional smoothing was done on the averaged curves: see \cite{github}.
\end{thebibliography}
\end{document}